\RequirePackage{amsmath}
\documentclass[runningheads, envcountsame, a4paper]{llncs}

\usepackage[utf8]{inputenc}
\usepackage[T1]{fontenc}
\usepackage[pdftex]{graphicx}
\usepackage{epstopdf}
\usepackage{microtype}

\usepackage[misc]{ifsym}

\usepackage{csquotes}
\usepackage[hyphens]{url}
\usepackage{acronym}

\usepackage{color}
\usepackage[colorlinks, allcolors=black,urlcolor=blue]{hyperref}

\usepackage[inline]{enumitem}

\usepackage{tabularx}

\usepackage{array}
\usepackage{multirow}
\usepackage{multicol}
\usepackage{booktabs}

\usepackage[caption=false]{subfig}

\usepackage{adjustbox}
\usepackage{nth}

\begin{document}
\title{Copy Mechanism and Tailored Training for Character-based Data-to-text Generation\thanks{Paper accepted to ECML-PKDD 2019. Please cite it using \href{https://dblp.uni-trier.de/rec/bib2/conf/pkdd/RobertiBCG19.bib}{these BibTeX entries}.}}
\toctitle{Copy Mechanism and Tailored Training for Character-based Data-to-text Generation}
\titlerunning{Copy Mechanism and Tailored Training for Character-based\dots}
%
\author{Marco~Roberti\inst{1}\textsuperscript{(\Letter)}
\and Giovanni~Bonetta\inst{1}
\and Rossella~Cancelliere\inst{1}
\and Patrick~Gallinari\inst{2,3}
}
\authorrunning{M. Roberti et al.}
%
\institute{University of Turin, Computer Science Department, Via Pessinetto, 12 -- 12149 Torino, Italy\\
\email{$\{$m.roberti, giovanni.bonetta, rossella.cancelliere$\}$@unito.it} \and
Sorbonne Universit\'e, 4 Place Jussieu -- 75005 Paris, France\\
\email{patrick.gallinari@lip6.fr} \and
Criteo AI Lab, 32 Rue Blanche -- 75009 Paris, France
}

\maketitle              
\setcounter{footnote}{0}
\tocauthor{Marco~Roberti \and Giovanni~Bonetta \and Rossella~Cancelliere \and Patrick~Gallinari}
\begin{abstract}
In the last few years, many different methods have been focusing on using deep recurrent neural networks for natural language generation. The most widely used sequence-to-sequence neural methods are word-based: as such, they need a pre-processing step called delexicalization (conversely, relexicalization) to deal with uncommon or unknown words. These forms of processing, however, give rise to models that depend on the vocabulary used and are not completely neural.

In this work, we present an end-to-end sequence-to-sequence model with attention mechanism which reads and generates at a {\it character level}, no longer requiring delexicalization, tokenization, nor even lowercasing. Moreover, since characters constitute the common ``building blocks" of every text, it also allows a more general approach to text generation,  enabling the possibility to exploit transfer learning for training.
These skills are obtained thanks to two major features: 
\begin{enumerate*}[label={(\roman*)}]
    \item the possibility to alternate between the standard generation mechanism and a copy one, which allows to directly copy input facts to produce outputs, and
    \item the use of an original training pipeline that further improves the quality of the generated texts.
\end{enumerate*}

We also introduce a new dataset called E2E+, designed to highlight the copying capabilities of character-based models, that is a modified version of the well-known E2E dataset used in the E2E Challenge. We tested our model according to five broadly accepted metrics (including the widely used \textsc{bleu}), showing that it yields competitive performance with respect to both character-based and word-based approaches.

\end{abstract}

\section{Introduction}
\label{introduction}

The ability of recurrent neural networks (RNNs) to model sequential data stimulated interest towards deep learning models which face data-to-text generation. An interesting application is the generation of descriptions for factual tables that consist of a set of field-value pairs; an example is shown in Table \ref{tab:outputs}. We present in this paper an effective end-to-end approach to this task.

Sequence-to-sequence frameworks~\cite{Cho:14,Sutskever:14,Aharoni:16} have proved to be very effective in natural language generation (NLG) tasks~\cite{Karpathy:14,Wen:15a,Mei:16}, as well as in machine translation~\cite{Cho:14,Sutskever:14,Bahdanau:14,Sennrich:16} and in language modeling~\cite{Al-Rfou:18}. Usually, data are represented word-by-word both in input and output sequences; anyways, such schemes can't be effective without a special, non-neural delexicalization phase that handles unknown words, such as proper names or foreign words (see ~\cite{Wen:15a}). 
The delexicalization step has the benefit of reducing the dictionary size and, consequently, the data sparsity, but it is affected by various shortcomings. In particular, according to~\cite{Goyal:16} - it needs some reliable mechanism for entity identification, i.e.\ the recognition of named entities inside text; - it requires a subsequent ``re-lexicalization'' phase, where the original named entities take back placeholders' place; - it cannot account for lexical or morphological variations due to the specific entity, such as gender and number agreements, that can't be achieved without a clear context awareness.

Recently, some strategies have been proposed to solve these issues:~\cite{Gu:16} and~\cite{See:17} face this problem using a special neural copying mechanism that is quite effective in alleviating the out-of-vocabulary words problem, while~\cite{Luong:15} tries to extend neural networks with a post-processing phase that copies words as indicated by the model's output sequence. Some character-level aspects appear as a solution of the issue as well, either as a fallback for rare words~\cite{Luong:16}, or as subword units~\cite{Sennrich:16}.

A significantly different approach consists in employing characters instead of words, for input slot-value pairs tokenization as well as for the generation of the final utterances, as done for instance in~\cite{Agarwal:17,Al-Rfou:18}.

In order to give an original contribution to the field, in this paper we present a character-level sequence-to-sequence model with attention mechanism that results in a completely neural end-to-end architecture. In contrast to traditional word-based ones, it does not require delexicalization, tokenization nor lowercasing; besides, according to our experiments it never hallucinates words, nor duplicates them. 
As we will see, such an approach achieves rather interesting performance results and produces a vocabulary-free model that is inherently more general, as it does not depend on a specific domain's set of terms, but rather on a general alphabet. 
Because of this, it opens up the possibility, not viable when using words, to adapt already trained networks to deal with different datasets.

More specifically, our model shows two important features, with respect to the state-of-art architecture proposed by~\cite{Bahdanau:14}:
\begin{enumerate*}[label={(\roman*)}]
    \item a character-wise copy mechanism, consisting in a soft switch between generation and copy mode, that disengages the model to learn rare and unhelpful self-correspondences, and
    \item a peculiar training procedure, which improves the internal representation capabilities, enhancing recall; it consists in the exchange of encoder and decoder RNNs, (GRUs~\cite{Cho:14} in our specific case), depending on whether the input is a tabular Meaning Representation (MR) or a natural language sentence.
\end{enumerate*}

As a further original contribution, we also introduce a new dataset, described in section \ref{sec:Datasets}, whose particular structure allows to better highlight improvements in copying/recalling abilities with respect to character-based state-of-art approaches.

In section \ref{sec:model}, after resuming the main ideas on encoder-decoder methods with attention, we detail our model: section \ref{sub:copy} is devoted to explaining the copy mechanism while in section \ref{sub:switch} our peculiar training procedure is presented. Section \ref{sec:experiments} includes the datasets descriptions, some implementation specifications, the experimental framework and the analysis and evaluation of the achieved results. Finally, in section \ref{sec:conclusion} some conclusions are drawn, outlining future work.

\begin{figure*}[t!]
    \centering
    \includegraphics[width=.75\textwidth]{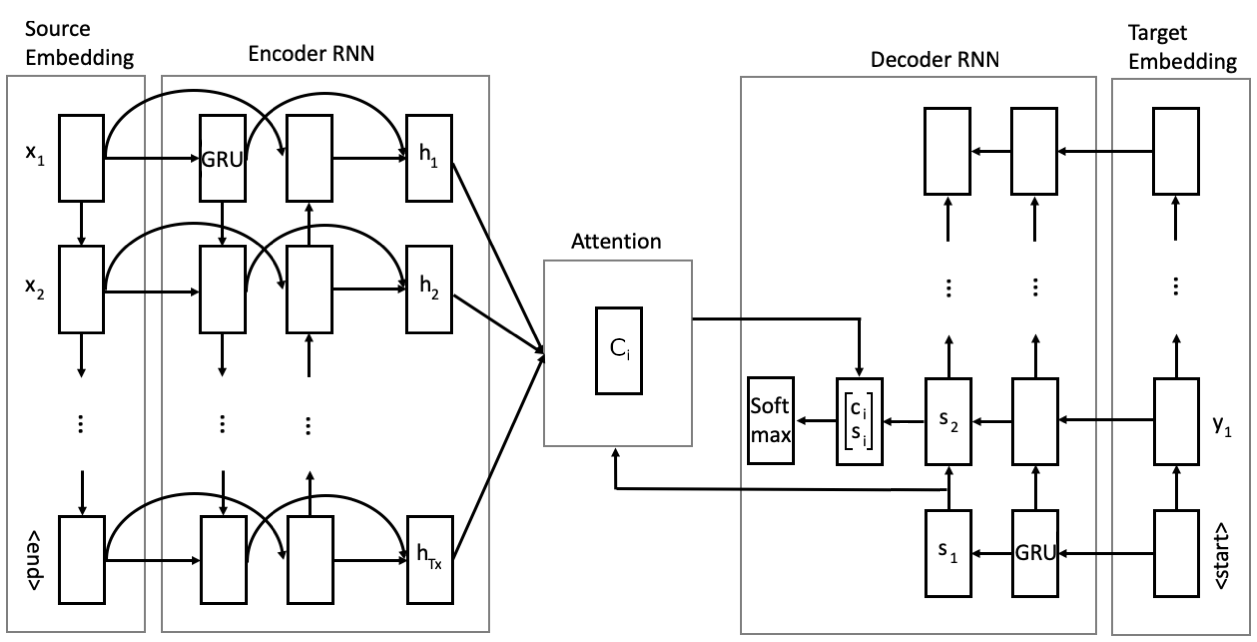}
    \caption{Encoder-decoder with attention model}
    \label{fig:baseline}
\end{figure*}

\section{Model Description}
\label{sec:model}
\subsection{Summary on Encoder-decoder Architectures with Attention }
\label{sub:bahdanau}

The sequence-to-sequence encoder-decoder architecture with attention~\cite{Bahdanau:14} is represented in figure \ref{fig:baseline}: on the left, the encoder, a bi-directional RNN, outputs one annotation $h_j$ for each input token $x_j$. Each vector $h_j$ corresponds to the concatenation of the hidden states produced by the backward and forward RNNs. On the right side of the figure, we find the decoder, which produces one state $s_i$ for each time step; on the center of the figure the attention mechanism is shown.
The main components of the attention mechanism are:

\noindent (i) the alignment model $e_{ij} $
\begin{equation}
    e_{ij} = att(s_{i-1},h_j),\: \: \: \: \: \:\: \: \: \quad 1\leq j \leq  T_x, \: \: \: 1\leq i \leq T_y
\end{equation}

\noindent which is parameterized as a feedforward neural network and scores how well input in position $j$-th and output observed in the $i$-th time instant match; $T_x$ and $T_y$ are the length of the input and output sequences, respectively. 
\\(ii) the attention probability distribution $  \alpha_{ij} $
\begin{equation}\label{eq:probatt}
    \alpha_{ij} = \frac{\exp(e_{ij})}{\sum_{k=1}^{T_x} \exp(e_{ik})}\equiv [softmax(e_i)]_j, \: \: \quad 1\leq j \leq  T_x, \: \: \: 1\leq i \leq T_y
\end{equation}
\noindent ($ e_i $ is the vector whose $j$-th element is $ e_{ij}$)
\\(iii) the context vector $ C_i $
\begin{equation}
    C_i = \sum_{j=1}^{T_x} \alpha_{ij}h_j,\: \: \: \quad 1\leq i \leq T_y,
\end{equation}
weighted sum of the encoder annotations $h_j$.  

According to~\cite{Bahdanau:14}, the context vector $ C_i$ is the key element for evaluating the conditional probability $ P(y_i|y_1, \dots, y_{i-1}, \textbf{x}) $ to output a target token $ y_i$, given the previously outputted tokens $ y_1, \dots, y_{i-1} $ and the input $ \textbf{x} $. They in fact express this probability as:
\begin{equation}\label{eq:probdec}
P(y_i|y_1, \dots, y_{i-1}, \textbf{x})=g(y_{i-1}, s_i, C_i),
\end{equation}
where $g$ is a non-linear, potentially multi-layered, function. So doing, the explicit information about $ y_1, \dots, y_{i-1} $ and $ \textbf{x} $ is replaced with the knowledge of the context $ C_i$ and the decoder state $ s_i$.

The model we present in this paper incorporates two additional mechanisms, detailed in the next sections: a character-wise copy mechanism and a peculiar training procedure based on GRUs switch.

\subsection{Learning to Copy}
\label{sub:copy}
On top of the just recalled model, we build a character-based copy mechanism inspired by the Pointer-Generator Network~\cite{See:17}, a word-based model that hybridizes the Bahdanau traditional model and a Pointer Network~\cite{Vinyals:15}. Basing on these ideas, in our model we identify two probability distributions that, differently from what done by~\cite{See:17} and~\cite{Wiseman:17},  {\it act now on characters} rather than on words: the alphabet distribution $ P_{alph} $ and the attention distribution $ P_{att}$. 

The former is the network's generative probability of sampling a given character at time $i$, recalled in eq. (\ref{eq:probdec}):
\begin{equation}
    P^i_{alph} = softmax(V[s_i;C_i] + b),
    \end{equation}
\noindent where $V$ and $b$ are trainable parameters.

The latter is the distribution reminded in eq. (\ref{eq:probatt}), created by the attention mechanism over the input tokens, i.e. in our case, over input characters:
\begin{equation}
    P_{att}^{ij} \equiv \alpha_{ij}
\end{equation}
In our method this distribution is used for directly copying characters from the input to the output, pointing their input positions, while in~\cite{Bahdanau:14} $P_{att}$ is used only internally to weigh the input annotations and create the context vector $C_i$. 

The final probability of outputting a specific character $ c $ is obtained combining $P_{alph}$ and $P_{att}$ through the quantity $p_{gen}$, defined later, which acts as a soft switch between generating $c$ or copying it:
\begin{equation}\label{eq:softswitch}
P^i(c) = p_{gen}^i \cdot P^i_{alph}[c] \\ + (1-p_{gen}^i)\sum_{j \mid x_i = c} P_{att}^{ij}(c),
\end{equation}
where $P^i_{alph}[c]$ is the component of $P^i_{alph}$ corresponding to that character $c$. 

The backpropagation training algorithm, therefore, brings $p_{gen}$ close to 1 when it is necessary to generate the output as in a standard encoder-decoder with attention ($P^{i}(c) \simeq P^i_{alph}[c]) $; conversely, $p_{gen}$ will be close to 0 (i.e.~$P^i(c) \simeq \sum_{j\mid x_i=c} P_{att}^j(c)$) when a copying step is needed.

The model we propose therefore learns when to sample from $P_{alph}$ for selecting the character to be generated, and when to sample from $P_{att}$ for selecting the character that has to be copied directly from the input. 

This copy mechanism is fundamental to output all the unknown words present in the input, i.e.~words which never occur in the training set. In fact, generating characters in the right order to reproduce unknown words is a sub-task not ``solvable'' by a naive sequence-to-sequence model, which learns to output only known words.


The generation probability $p_{gen} \in [0,1]$ is computed as follows:
\begin{equation}
p_{gen}^i = \sigma(W_y\cdot\tilde{y}_{i-1} + W_s \cdot s_{i} \\+ W_p \cdot p_{gen}^{i-1} + W_c \cdot C_i)
\end{equation}
where $\sigma$ is the sigmoid function, $\tilde{y}_{i-1}$ is the last output character's embedding, $s_i$ is the current decoder's cell state and $C_i$ is the current context vector. $W_y$, $W_s$, $W_c$ and $W_p$ are the parameters whose training allows $p_{gen} $ to have the convenient value.

We highlight that in our formulation $p_{gen}^{i-1}$, i.e. the value of $p_{gen}$ at time $i-1$, contributes to the determination of $p_{gen}^i$. In fact, in a character-based model it is desirable that this probability remains unchanged for a fair number of time steps, and knowing its last value helps this behavior. This never happens in word-based models (such as~\cite{See:17}), in which copying for a single time step is usually enough.

\subsection{Switching GRUs}
\label{sub:switch}
Aiming at improving performance, we enrich our model' training pipeline with an additional phase, which forces an appropriate language representation inside the recurrent components of the model. In order to achieve this goal, the encoder and the decoder {\it do not own a fixed GRU}, differently from what happens in classical end-to-end approaches. The recurrent module is passed each time as a parameter, depending on which one of the two training phases is actually performed.

In the first phase, similar to the usual one, the GRU assigned to the encoder deals with a tabular representation $x$ as input, the GRU assigned to the decoder has to cope with natural language, and the model generates an output utterance $\tilde{y} = F(x)$. Conversely, in the second phase GRUs are switched and we use as input the just obtained natural language utterance $ \tilde{y} $ to generate a new table $\tilde{x}=G(\tilde{y}) = G(F(x))$. Therefore, the same model can build both $F$ and $G$, thanks to the switch of GRUs.


In other words, the learning iteration is performed as follows.
\begin{itemize}
    \item A dataset example $(x, y)$ is given. $x$ is a tabular meaning representation and $y$ is the corresponding reference sentence.
    \item We generate an output utterance $ \tilde{y} = F(x) $ 
    \item We perform an optimization step on the model's parameters, aiming at minimizing $ L_{forward} = loss(\tilde{y}, y) $ 
    \item We reconstruct the meaning representation $\tilde{x}$ back from the previously generated output: $ \tilde{x} = G(\tilde{y}) = G(F(x)) $
    \item We perform a further optimization step on the model's parameters, this time aiming at minimizing $ L_{backward} = loss(\tilde{x}, x) $ 
\end{itemize}

The higher training time, direct consequence of the just described technique, is a convenient investment, as it brings an appreciable improvement of the model's performance (see section \ref{sub:results}).


\section{Experiments}
\label{sec:experiments}

\subsection{Datasets}
\label{sec:Datasets}

\begin{table*}[t!]
\caption{\label{tab:datasets} Descriptive statistics: on the left, sizes of training, validation and test sets are shown. On the right, the average number of characters, respectively for Meaning Representations and natural language sentences, are presented}
\centering
\begin{tabular}{lcccc>{\centering\arraybackslash}p{1.5cm}c}
  \toprule
  \multirow{2}*{Dataset} & \multicolumn{3}{c}{Number of instances} && \multicolumn{2}{c}{Avg. number of characters} \\
  \cmidrule(rl){2-4}\cmidrule(rl){6-7}
  & training & validation & test && MRs & NL sentences \\
  \midrule
  E2E        & 42061 & 4672 & 4693 && 112.11 & 115.07 \\
  E2E+       & 42061 & 4672 & 4693 && 112.91 & 115.65 \\
  Hotel      &  2210 &  275 &  275 &&  52.74 &  61.31 \\
  Restaurant &  2874 &  358 &  358 &\qquad\quad&  53.89 &  63.22 \\
  \bottomrule
\end{tabular}

\end{table*}

We tested our model on four datasets, whose main descriptive statistics are given in table \ref{tab:datasets}: among them, the most known and frequently used in literature is the E2E dataset~\cite{Novikova:17}, used as benchmark for the E2E Challenge organized by the Heriot-Watt University in 2017. It is a crowdsourced collection of roughly 50,000 instances, in which every input is a list of slot-value pairs and every expected output is the corresponding natural language sentence. The dataset has been partitioned by the challenge organizers in predefined training, validation and test sets, conceived for training data-driven, end-to-end Natural Language Generation models in the restaurant domain.

However, during our experiments, we noticed that the values contained in the E2E dataset are a little naive in terms of variability. In other words, a slot like \textit{name}, that could virtually contain a very broad range of different values, is filled alternating between 19 fixed possibilities. Moreover, values are partitioned among training, validation and test set, in such a way that test set always contains values that are also present in the training set.
Consequently, we created a modified version of the E2E dataset, called E2E+, as follows: we selected the slots that represent more copy-susceptible attributes, i.e. \textit{name}, \textit{near} and \textit{food}, and conveniently replaced their values, in both meaning representations and reference sentences. New values for \textit{food} are picked from Wikipedia's list of adjectival forms of countries and nations\footnote{\url{https://en.wikipedia.org/wiki/List_of_adjectival_and_demonymic_forms_for_countries_and_nations}, consulted on August 30, 2018}, while both \textit{name} and \textit{near} are filled with New York restaurants' names contained in the Entree dataset presented in~\cite{Burke:97}. It is worth noting that none of the values of \textit{name} are found in \textit{near}; likewise, values that belong to the training set are not found in the validation set nor in the test one, and vice versa. This value partitioning shall ensure the absence of generation bias in the copy mechanism, stimulating the models to copy attribute values, regardless of their presence in the training set. The \textit{MR} and \textit{1st reference} fields in table \ref{tab:outputs} are instances of this new dataset.

Finally, we decided to test our model also on two datasets, Hotel and Restaurant, frequently used in literature (for instance in~\cite{Wen:15a} and~\cite{Goyal:16}). They are built on a 12 attributes ontology: some attributes are common to both domains, while others are domain specific. Every MR is a list of key-value pairs enclosed in a dialogue act type, such as \textit{inform}, used to present information about restaurants, \textit{confirm}, to check that a slot value has been recognized correctly, and \textit{reject}, to advise that the user's constraints cannot be met. For the sake of compatibility, we filtered out from Hotel and Restaurant all inputs whose dialogue act type was not \textit{inform}, and removed the dialogue act type. Besides, we changed the format of the key-value pairs to E2E-like ones.

Tables are encoded simply converting all characters to ASCII and feeding every corresponding index to the encoder, sequentially. The resulting model's vocabulary is independent of the input, allowing the application of the transfer learning procedure.

\subsection{Implementation Details}
We developed our system using the PyTorch framework\footnote{Code and datasets are publicly available at \url{https://github.com/marco-roberti/char-data-to-text-gen}}, release 0.4.1\footnote{\url{https://pytorch.org/}}.
The training has been carried out as described in subsection \ref{sub:switch}: this training procedure needs the two GRUs to have the same dimensions, in terms of input size, hidden size, number of layers and presence of a bias term. Moreover, they both have to be bidirectional, even if the decoder ignores the backward part of its current GRU. 

We minimize the negative log-likelihood loss using teacher forcing~\cite{Williams:89} and Adam~\cite{Kingma:14}, the latter being an optimizer that computes individual adaptive learning rates. As a consequence of the length of the input sequences, a character-based model is often subject to the exploding gradient problem, that we solved via the well-known technique of gradient norm clipping~\cite{Pascanu:13}.

We also propose a new formulation of $P(c)$ that helps the model to learn when it is necessary to start a copying phase:

\begin{equation}\label{eq:shift}
P^i(c) = p_{gen}^i \cdot P_{alph}^i(c) \\+ (1-p_{gen}^i)\sum_{j \mid x_i = c} P_{att}^{i,j-1}(c)
\end{equation}

Sometimes, our model has difficulty in focusing on the first letter it has to copy. This may be caused by the variety of characters it could be attending on; instead, it seems easier to learn to focus on the most largely seen characters, as for instance ` ' and `['. As these special characters are very often the prefix of the words we need to copy, when this focus is achieved, we would like the attention distribution to be translated one step to the right, over the first letter that must be copied. Therefore, the final probability of outputting a specific character $c$, introduced in eq.~(\ref{eq:softswitch}), is modified to $P_{att}^{i,j-1}$, i.e.~the attention distribution shifted one step to the right and normalized.

Notice that $P_{att}^{i,j-1}$ is the only shifted probability, while $P_{alph}^i$ remains unchanged. Therefore, if the network is generating the next token (i.e. $p_{gen}^i \simeq 1$ ), the shift trick does not involve $P^i(c)$ and the network samples the next character from $P_{alph}^i$, as usual. This means that the shift operation is not degrading the generation ability of the model, whilst improving the copying one.

\subsection{Results and Discussion}
\label{sub:results}
In order to show that our model represents an effective and relevant improvement, we carry out two different experimentations: an ablation study and a comparison with two well-known models. The first model is the encoder-decoder architecture with attention mechanism by~\cite{Bahdanau:14} (hereafter ``EDA''), used character-by-character. The second one is TGen~\cite{Dusek:16}, a word-based model, still derived from~\cite{Bahdanau:14}, but integrating a beam search mechanism and a reranker over the top $k$ outputs, in order to disadvantage utterances that do not verbalize all the information contained in the MR.  We chose it because it has been adopted as baseline in the E2E NLG Challenge\footnote{\url{www.macs.hw.ac.uk/InteractionLab/E2E/}}.

We used the official code provided in the E2E NLG Challenge website for TGen, and we developed our models and EDA in PyTorch, training them on NVIDIA GPUs. Hyperparameter tuning is done through 10-fold cross-validation, using the \textsc{bleu} metric~\cite{Papineni:02} for evaluating each model. The training stopping criterion was based on the absence of models' performance improvements (see~\cite{Dusek:16}).

\begin{table*}[t!]
\caption{\label{tab:ablation} The ablation study on the E2E dataset evidences the final performance improvement reached by our model. Best values for each metric are highlighted (the higher the better)}

\centering

\subfloat{
\begin{tabular}{llcccc}
  \toprule
  \multirow{5}*{EDA}     & \textsc{bleu}     & 0.4999 \\
                         & \textsc{nist}     & 7.1146 \\
                         & \textsc{meteor}   & 0.3369 \\
                         & \textsc{rouge\_l} & 0.5634 \\
                         & \textsc{cider}    & 1.3176 \\
  \midrule
  \multirow{5}*{EDA\_C}  & \textsc{bleu}     & 0.6255 \\
                         & \textsc{nist}     & 7.7934 \\
                         & \textsc{meteor}   & 0.4401 \\
                         & \textsc{rouge\_l} & 0.6582 \\
                         & \textsc{cider}    & 1.7286 \\
  \bottomrule
\end{tabular}
} \qquad\qquad
\subfloat{
\begin{tabular}{llcccc}
  \toprule
  \multirow{5}*{EDA\_S}  & \textsc{bleu}     & 0.6538 \\
                         & \textsc{nist}     & 8.4601 \\
                         & \textsc{meteor}   & 0.4337 \\
                         & \textsc{rouge\_l} & 0.6646 \\
                         & \textsc{cider}    & 1.9944 \\
  \midrule
  \multirow{5}*{EDA\_CS} & \textsc{bleu}     & \bf 0.6705 \\
                         & \textsc{nist}     & \bf 8.5150 \\
                         & \textsc{meteor}   & \bf 0.4449 \\
                         & \textsc{rouge\_l} & \bf 0.6894 \\
                         & \textsc{cider}    & \bf 2.2355 \\
  \bottomrule
\end{tabular}
}

\end{table*}


We evaluated the models' performance on test sets' output utterances using the Evaluation metrics script\footnote{\url{https://github.com/tuetschek/E2E-metrics}} provided by the E2E NLG Challenge organizers. It rates quality according to five different metrics: \textsc{bleu}~\cite{Papineni:02}, \textsc{nist}~\cite{Doddington:02}, \textsc{meteor}~\cite{Lavie:07}, \textsc{rouge\_l}~\cite{Lin:04} and \textsc{cider}~\cite{Vedantam:15}.

Our first experimentation, the \textbf{ablation study}, refers to the E2E dataset because of its wide diffusion, and is shown in table \ref{tab:ablation}; ``EDA\_CS'' identifies our model, and `C' and `S' stand for ``Copy'' and ``Switch'', the two major improvements presented in this work. It is evident that the partially-improved networks are able to provide independent benefits to the performance. Those components cooperate positively, as EDA\_CS further enhances those results. Furthermore, the obtained \textsc{bleu} metric value on the E2E test set would allow our model to be ranked fourth in the E2E NLG Challenge, while its baseline TGen was ranked tenth. 

\begin{table*}[t!]
\caption{\label{tab:transfer} Performance comparison. Note the absence of transfer learning on dataset E2E+ because in this case the training and fine-tuning datasets are the same. Best values for each metric are highlighted (the higher the better)}
\centering
\begin{tabular}{llcccc}
  \toprule
                             &                   & E2E+   & E2E    & Hotel  & Restaurant \\
  \midrule
  \multirow{5}*{EDA}         & \textsc{bleu}     & 0.3773 & 0.4999 & 0.4316 & 0.3599 \\
                             & \textsc{nist}     & 5.7835 & 7.1146 & 5.9708 & 5.5104 \\
                             & \textsc{meteor}   & 0.2672 & 0.3369 & 0.3552 & 0.3367 \\
                             & \textsc{rouge\_l} & 0.4638 & 0.5634 & 0.6609 & 0.5892 \\
                             & \textsc{cider}    & 0.2689 & 1.3176 & 3.9213 & 3.3792 \\
  \midrule
  \multirow{5}*{TGen}        & \textsc{bleu}     & \bf 0.6292 & 0.6593 & 0.5059 & 0.4074 \\
                             & \textsc{nist}     & \bf 9.4070 & \bf 8.6094 & 7.0913 & 6.4304 \\
                             & \textsc{meteor}   & 0.4367 & 0.4483 & 0.4246 & 0.3760 \\
                             & \textsc{rouge\_l} & \bf 0.6724 & 0.6850 & 0.7277 & 0.6395 \\
                             & \textsc{cider}    & 2.8004 & 2.2338 & 5.0404 & 4.1650 \\
  \midrule
  \multirow{5}*{EDA\_CS}     & \textsc{bleu}     & 0.6197 & \bf 0.6705 & 0.5515 & 0.4925 \\
                             & \textsc{nist}     & 9.2103 & 8.5150 & \bf 7.4447 & 6.9813 \\
                             & \textsc{meteor}   & \bf 0.4428 & 0.4449 & 0.4379 & 0.4191 \\
                             & \textsc{rouge\_l} & 0.6610 & \bf 0.6894 & 0.7499 & 0.7002 \\
                             & \textsc{cider}    & \bf 2.8118 & \bf 2.2355 & 5.1376 & 4.7821 \\
  \midrule
  \multirow{5}*{EDA\_CS\textsuperscript{\it TL}} & \textsc{bleu}     &  -  & 0.6580 & \bf 0.5769 & \bf 0.5099 \\
                             & \textsc{nist}     &  -  & 8.5615 & 7.4286 & \bf 7.3359 \\
                             & \textsc{meteor}   &  -  & \bf 0.4516 & \bf 0.4439 & \bf 0.4340 \\
                             & \textsc{rouge\_l} &  -  & 0.6740 & \bf 0.7616 & \bf 0.7131 \\
                             & \textsc{cider}    &  -  & 2.1803 & \bf 5.3456 & \bf 4.9915 \\
  \bottomrule
\end{tabular}
\end{table*}

Our second experimentation, the \textbf{comparison study}, is shown in table \ref{tab:transfer}. The character-based design of EDA\_CS led us to explore in this context also a possible behavior as a transfer learning capable model: in order to test this hypothesis, we used the weights learned during training on the E2E+ dataset as the starting point for a fine-tuning phase on all the other datasets. We chose E2E+ because it reduces the generation bias, as discussed in subsection \ref{sec:Datasets}. We named this approach EDA\_CS\textsuperscript{\it TL}.

A first interesting result is that our model EDA\_CS always obtains higher metric values with respect to TGen on the Hotel and Restaurant datasets, and three out of five higher metrics values on the E2E dataset. However, in the case of E2E+, TGen achieves three out of five higher metrics values. These results suggest that EDA\_CS and TGen are comparable, at least from the point of view of automatic metrics' evaluation.

A more surprising result is that the approach EDA\_CS\textsuperscript{\it TL} allows to obtain better performance with respect to training EDA\_CS in the standard way on the Hotel and Restaurant datasets (for the majority of metrics); on E2E, EDA\_CS\textsuperscript{\it TL} outperforms EDA\_CS only in one case (i.e. \textsc{meteor} metric).

Moreover, EDA\_CS\textsuperscript{\it TL} shows a \textsc{bleu} increment of at least $14\%$ with respect to TGen's score when compared to both Hotel and Restaurant datasets.

Finally, the baseline model, EDA, is largely outperformed by all other examined methods.

Therefore, we can claim that our model exploits its transfer learning capabilities effectively, showing very good performances in a context like data-to-text generation in which the portability of features learned from different datasets, in the extent of our knowledge, has not yet been explored.



We highlight that EDA\_CS's model's good results are achieved even if it consists in a fully end-to-end model which does not benefit from the delexicalization-relexicalization procedure, differently from TGen. Most importantly, the latter represents a word-based system: as such, it is bound to a specific, limited vocabulary, in contrast to the general-purpose character one used in our work.

\begin{table*}[t!]
\caption[A comparison of the three models' output]{\label{tab:outputs} A comparison of the three models' output on some MR of the E2E+ test set. The first reference utterance is reported for convenience}
\begin{adjustbox}{center}
\footnotesize
\begin{tabularx}{\textwidth}{p{0.2\textwidth}X}
  \toprule
  \multirow{2}*{MR} & \texttt{name[New Viet Huong], eatType[pub], customer rating[1 out of 5], near[Ecco]} \\
  \multirow{2}*{\nth{1} reference} & The New Viet Huong is a pub near Ecco that has a customer rating of 1 out of 5. \\
  \multirow{2}*{EDA\_CS} & New Viet Huong is a pub near Ecco with a customer rating of 1 out of 5. \\
  \multirow{2}*{TGen} & New Viet Huong is a pub near Ecco with a customer rating of 1 out of 5. \\
  \multirow{2}*{EDA} & Near the riverside near the ERNick Restaurant is a pub near the ERNicker's. \\
  \midrule
  \multirow{3}*{MR} & \texttt{name[La Mirabelle], eatType[restaurant], food[Iraqi], priceRange[high], area[riverside], familyFriendly[yes], near[Mi Cocina]} \\
  \multirow{3}*{\nth{1} reference} & La Mirabelle is a children friendly restaurant located in the Riverside area near to the Mi Cocina. It serves Iraqi food and is in the high price range. \\
  \multirow{2}*{EDA\_CS} & La Mirabelle is a high priced Iraqi restaurant located in the riverside area near Mi Cocina. It is children friendly. \\
  \multirow{2}*{TGen} & La Mirabelle is a high priced Iraqi restaurant in the riverside area near Mi Cocina. It is child friendly. \\
  \multirow{3}*{EDA} & La Memaini is a high priced restaurant that serves Iranian food in the high price range. It is located in the riverside area near Manganaro's Restaurant. \\
  \bottomrule
\end{tabularx}
\end{adjustbox}
\end{table*}

Table \ref{tab:outputs} reports the output of the analyzed models for a couple of MR, taken from the E2E+ test set. The EDA's inability to copy is clear, as it tends, in its output, to substitute those values of \textit{name}, \textit{food} and \textit{near} that do not appear in the training set with known ones, guided by the first few characters of the input slot's content. Besides, it shows serious coverage issues, frequently 'forgetting' to report information, and/or repeating more times the same ones. 

 These troubles are not present in EDA\_CS output utterances: the model nearly always renders all of the input slots, still without duplicating any of them. This goal is achieved even in absence of explicit coverage techniques thanks to our peculiar training procedure, detailed in section \ref{sub:switch}, that for each input sample minimizes also the loss on the reconstructed tabular input. It is worth noting that the performance of TGen and EDA\_CS are overall comparable, especially when they deal with names or other expressions not present in training. 

\begin{figure}
    \centering
    \subfloat[\emph{On an E2E instance}.\label{subfig:E2E}]
    {\includegraphics[width=.685\textwidth]{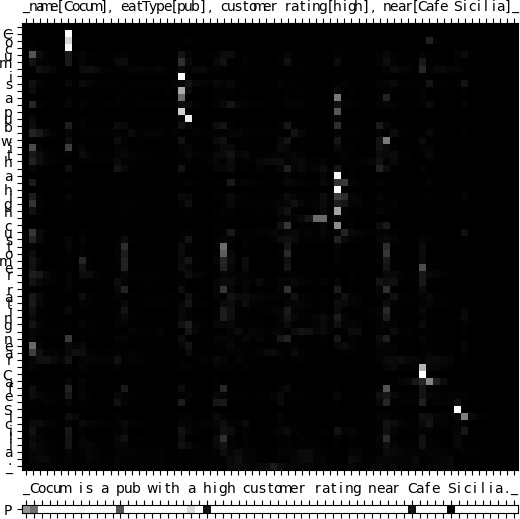}} \\
    \subfloat[\emph{On an E2E+ instance}.\label{subfig:E2E+}]
    {\includegraphics[width=.735\textwidth]{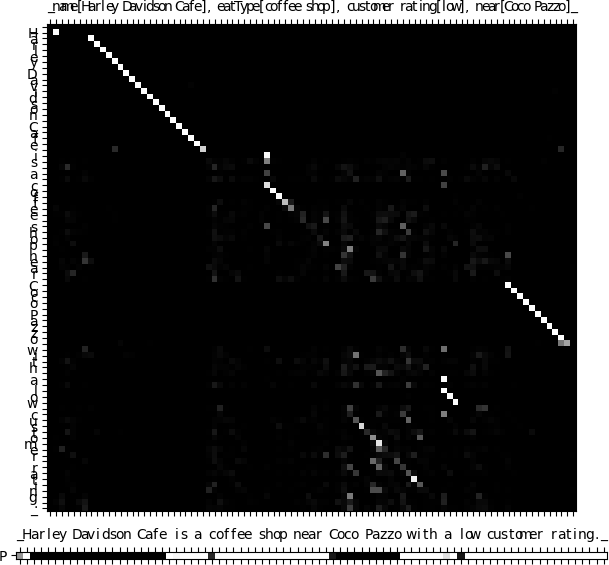}}

    \caption[Attention and $p_{gen}$]{Attention distribution (white means more attention) and $p_{gen}$ (white: generating, black: copying), as calculated by the model}
    \label{fig:attention}
\end{figure}

\begin{figure}
    \centering
    \includegraphics[width=.45\textwidth]{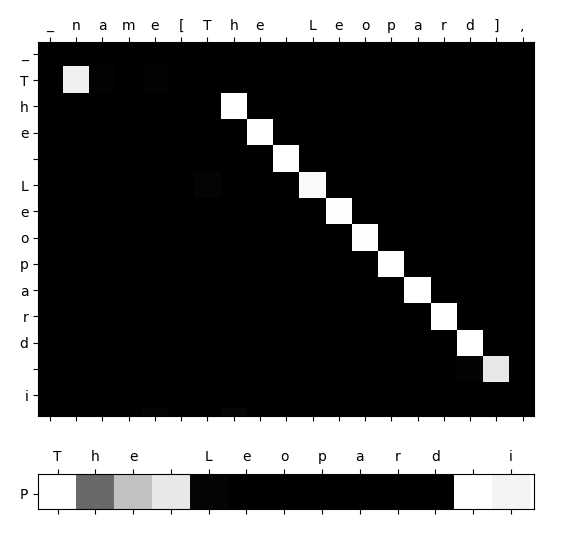}
    \caption{Copying common words leads the model to ``uncertain'' values of $p_{gen}$}
    \label{fig:common}
\end{figure}

The joint analysis of the matrix of the attention distribution $P^{ij}_{att}$ and the vector $p_{gen}$ allows a deeper understanding of how our model works. 

In figure \ref{fig:attention} every row shows the attention probability distribution ``seen'' when an output character is produced at the $i$-th time instant (i.e. the vector $ P^{ij}_{att}, 1\leq j\leq T_x $), while every column shows values of the attention distribution corresponding to a specific input position $ j $ (i.e. the vector $ P^{ij}_{att}, 1 \leq i \leq T_y $). We can therefore follow the white spots, corresponding to higher values of attention, to understand the flow of the model's attention during the generation of the output utterance. 

Moreover, $p_{gen}$ values, which lie in the numeric interval $[0,1]$, help us in the interpretation of the attention: they are represented as a grayscale vector from zero (black) to one (white) under the matrices. Values close to $0$ mean copying and those near $1$ mean generating.

We can note that our model's behavior varies significantly depending on the dataset it has been trained on. Figure \ref{subfig:E2E} shows the attention probability distribution matrix of EDA\_CS (together with $p_{gen}$ vector) trained on the E2E dataset: as observed before, attribute values in this dataset have a very low variability (and are already present in the training set), so that they can be individually represented and easily generated by the decoder. In this case, a typical pattern is the copy of only the first, discriminating character, clearly noticeable in the graphical representation of the $p_{gen}$ vector, and the subsequent generation of the others. Notice that the attention tends to remain improperly focused on the same character for more than one output time step, as in the first letter of ``high''.

On the other hand, the copy mechanism shows its full potential when the system must learn to copy attribute values, as in the E2E+ dataset. In figure \ref{subfig:E2E+} the diagonal attention pattern is pervasive:
\begin{enumerate*}[label={(\roman*)}]
    \item it occurs when the model actually copies, as in ``Harley Davidson'' and ``Coco Pazzo'', and 
    \item as a \textit{soft track} for the generation, as in ``customer rating'', where the copy-first-generate-rest behavior emerges again.
\end{enumerate*}

A surprising effect is shown in figure \ref{fig:common}, when the model is expected to copy words that, instead, are usually generated: an initial difficulty in copying the word ``The'', that is usually a substring of a slot value, is ingeniously overcome as follows. The first character is purely generated, as shown by the white color in the underlying vector, and the sequence of the following characters, ``he\_'', is half-generated and half-copied. Then, the value of $p_{gen}$ gets suddenly but correctly close to $0$ (black) until the closing square bracket is met. The network's output is not affected negatively by this confusion and the attention matrix remains quite well-formed.

As a final remark, the metrics used, while being useful, well-known and broadly accepted, do not reflect the ability to directly copy input facts to produce outputs, so settling the rare word problem.

\section{Conclusion}
\label{sec:conclusion}
We showed in this paper an effective character-based end-to-end model that faces data-to-text generation tasks. It takes advantage of a copy mechanism, that deals successfully with the rare word problem, and of a specific training procedure, characterized by the switching GRUs mechanism. These innovative contributions to state-of-art further improve the quality of the generated texts.

We highlight that our formulation of the copy mechanism is an original character-based adaptation of~\cite{See:17}, because of the use of $p_{gen}^{i-1}$ to determine the value of $p_{gen}^i$, at the following time step. This helps the model in choosing whether to maintain the same value for a fair number of time steps or not.

Besides, the use of characters allows the creation of more general models, which do not depend on a specific vocabulary; it also enables a very effective straightforward transfer learning procedure, which in addition eases training on small datasets. Moreover, outputs are obtained in a completely end-to-end fashion, in contrast to what happens for the chosen baseline word-based model, whose performances are comparable or even worse.

One future improvement of our model could be the ``reinforcement'' of the learning iteration described in section \ref{sub:switch}: for each dataset example $(x,y)$, we could consider, as an ulterior example, the reverse instance $(y,x)$. The network obtained this way should be completely reversible, and the interchangeability of input and output languages could open up new opportunities in neural machine translation, such as two-way neural translators.

New metrics that give greater importance to rare words might be needed in the future, with the purpose of better assess performances of able-to-copy NLG models on datasets such as the E2E+ one.

\section*{Acknowledgements}
The activity has been partially carried on in the context of the Visiting Professor Program of the Italian Istituto Nazionale di Alta Matematica (INdAM).


\end{document}